\documentclass[11pt,a4paper]{article}
\usepackage[hyperref]{emnlp2018}
\usepackage{times}
\usepackage{latexsym}
\usepackage{multicol}
\usepackage{multirow}

\usepackage{url}

\usepackage{graphicx}
\usepackage{caption}
\usepackage{booktabs}
\usepackage{enumitem}
\usepackage{xspace}
\usepackage{multirow}

\usepackage{amsmath}
\usepackage{amsfonts}
\usepackage{amsthm, url}

\usepackage{todonotes}
\usepackage{lipsum}
\usepackage{slashbox}

\usepackage{subcaption}

\aclfinalcopy 
\def\aclpaperid{628} 

\hypersetup{draft}

\def\name{CLOTH\xspace}
\def\namem{CLOTH-M\xspace}
\def\nameh{CLOTH-H\xspace}

\title{Large-scale Cloze Test Dataset Created by Teachers}

\author{Qizhe Xie\thanks{~~~Equal contribution.}~~, Guokun Lai\footnotemark[1]~~, Zihang Dai, Eduard Hovy \\
  Language Technologies Institute, Carnegie Melon University \\
  {\tt \{qizhex, guokun, dzihang, hovy\}@cs.cmu.edu}
  }
\date{}

\begin{document}
\maketitle
\begin{abstract}
Cloze tests are widely adopted in language exams to evaluate students' language proficiency. In this paper, we propose the first large-scale human-created cloze test dataset CLOTH
\footnote{CLOTH (\textbf{CLO}ze test by \textbf{T}eac\textbf{H}ers) is available at \url{http://www.cs.cmu.edu/~glai1/data/cloth/}.} \footnote{The leaderboard is available at \url{http://www.qizhexie.com/data/CLOTH_leaderboard.html}}, 
containing questions used in middle-school and high-school language exams. With missing blanks carefully created by teachers and candidate choices purposely designed to be nuanced, CLOTH requires a deeper language understanding and a wider attention span than previously automatically-generated cloze datasets. We test the performance of dedicatedly designed baseline models including a language model trained on the One Billion Word Corpus and show humans outperform them by a significant margin. We investigate the source of the performance gap, trace model deficiencies to some distinct properties of CLOTH, and identify the limited ability of comprehending the long-term context to be the key bottleneck. 
\end{abstract}
\section{Introduction}
Being a classic language exercise, the cloze test~\citep{taylor1953cloze} is an accurate assessment of language proficiency~\citep{fotos1991cloze, jonz1991cloze, tremblay2011proficiency} and has been widely employed in language examinations. 
Under a typical setting, a cloze test requires examinees to fill in missing words (or sentences) to best fit the surrounding context. 
To facilitate natural language understanding, automatically-generated cloze datasets are introduced to measure the ability of machines in reading comprehension~\citep{hermann2015teaching, hill2015goldilocks, onishi2016did}. 
In these datasets, each cloze question typically consists of a context paragraph and a question sentence.
By randomly replacing a particular word in the question sentence with a blank symbol, a single test case is created.
For instance, CNN/Daily Mail datasets~\citep{hermann2015teaching} use news articles as contexts and summary bullet points as the question sentence. Only named entities are removed when creating the blanks.
Similarly, in Children's Books test (CBT)~\citep{hill2015goldilocks},  cloze questions are obtained by removing a word in the last sentence of every consecutive $21$ sentences, with the first 20 sentences being the context.
Different from CNN/Daily Mail datasets, CBT also provides each question with a candidate answer set, consisting of randomly sampled words with the same part-of-speech tag from the context as that of the correct answer.

Thanks to the automatic generation process, these datasets can be very large in size, leading to significant research progresses. However, compared to how humans would create cloze questions and evaluate reading comprehension ability, the automatic generation process bears some inevitable issues. 
Firstly, blanks are chosen uniformly without considering which aspect of the language phenomenon that questions will test. Hence, quite a portion of automatically-generated questions can be purposeless or even trivial to answer.
Another issue involves the ambiguity of answers. Given a context and a sentence with a blank, there can be multiple words that fit almost equally well into the blank. A possible solution is to include a candidate option set, as done by CBT, to get rid of the ambiguity. However, automatically generating the candidate option set can be problematic since it cannot guarantee the ambiguity is removed. More importantly, automatically-generated candidates can be totally irrelevant or simply grammatically unsuitable for the blank, resulting in again purposeless or trivial questions.
Probably due to these unsatisfactory issues, neural models have achieved comparable results to the human-level performance within a very short time~\citep{chen2016thorough,dhingra2016gated, seo2016bidirectional}.
While there have been works trying to incorporate human design into cloze question generation~\citep{zweig2011microsoft, paperno2016lambada}, due to the expensive labeling process, the MSR Sentence Completion Challenge created by this effort has $1,040$ questions and the LAMBADA~\citep{paperno2016lambada} dataset has $10,022$ questions, limiting the possibility of developing powerful neural models on it. As a result of the small size, human-created questions are only used to compose development sets and test sets. 
Motivated by the aforementioned drawbacks, we propose \name, a large-scale cloze test dataset collected from English exams. Questions in the dataset are designed by middle-school and high-school teachers to prepare Chinese students for entrance exams. To design a cloze test, teachers firstly determine the words that can test students' knowledge of vocabulary, reasoning or grammar; 
then replace those words with blanks and provide other three candidate options for each blank. If a question does not specifically test grammar usage, all of the candidate options would complete the sentence with correct grammar, leading to highly nuanced questions. As a result, human-created questions are usually harder and are a better assessment of language proficiency. A general cloze test evaluates several aspects of language proficiency including vocabulary, reasoning and grammar, which are key components of comprehending natural language. 

To verify if human-created cloze questions are difficult for current models, we train and evaluate the state-of-the-art language model (LM) and machine comprehension models on this dataset, including a language model trained on the One Billion Word Corpus. We find that the state-of-the-art model lags behind human performance even if the model is trained on a large external corpus. We analyze where the model fails compared to humans who perform well. After conducting error analysis, we assume the performance gap results from the model's inability to use a long-term context. To examine this assumption, we evaluate human-level performance when the human subjects are only allowed to see one sentence as the context. Our assumption is confirmed by the matched performances of the models and human when given only one sentence. 
In addition, we demonstrate that human-created data is more difficult than automatically-generated data. Specifically, it is much easier for the same model to perform well on automatically-generated data. 

We hope that \name provides a valuable testbed for both the language modeling community and the machine comprehension community. Specifically, the language modeling community can use \name to evaluate their models' abilities in modeling long contexts, while the machine comprehension community can use \name to test machine's understanding of language phenomena.
\section{Related Work}
Large-scale automatically-generated cloze tests~\citep{hermann2015teaching, hill2015goldilocks,onishi2016did} lead to significant research advancements. However, generated questions do not consider language phenomenon to be tested and are relatively easy to solve. Recently proposed reading comprehension datasets are all labeled by humans to ensure a high quality~\citep{rajpurkar2016squad,joshi2017triviaqa,trischler2016newsqa,nguyen2016ms}. 

Perhaps the closet work to \name is the LAMBADA dataset ~\citep{paperno2016lambada}. 
LAMBADA also targets at finding challenging words to test LM's ability in comprehending a longer context. However, LAMBADA does not provide a candidate set for each question, which can cause ambiguities when multiple words can fit in. Furthermore, only test set and development set are labeled manually. The provided training set is the unlabeled Book Corpus~\citep{zhu2015aligning}. Such unlabeled data do not emphasize long-dependency questions and have a mismatched distribution with the test set, as showed in Section \ref{sec:difficulty}. Further, the Book Corpus is too large to allow rapid algorithm development for researchers who do not have access to a huge amount of computational power. 

Aiming to evaluate machines under the same conditions that the humans are evaluated, there is a growing interest in obtaining data from examinations. 
NTCIR QA Lab~\citep{shibuki2014overview} contains a set of real-world college entrance exam questions. The Entrance Exams task at CLEF QA Track~\citep{penas2014overview, rodrigo2015overview} evaluates machine's reading comprehension ability. The AI2 Reasoning Challenge~\citep{clark2018think, schoenick2017moving} contains approximately eight thousand scientific questions used in middle school. 
\citet{lai2017large} proposes the first large-scale machine comprehension dataset obtained from exams. They show that questions designed by teachers have a significantly larger proportion of reasoning questions. Our dataset focuses on evaluating both language proficiency and reasoning abilities. 

\section{\name Dataset}
In this section, we introduce the \name dataset that is collected from English examinations, and study its abilities of assessment. 
\subsection{Data Collection and Statistics}
We collect the raw data from three free and public websites in China that gather exams created by English teachers to prepare students for college/high school entrance exams\footnote{
The three websites include http://www.21cnjy.com/; http://5utk.ks5u.com/; http://zujuan.xkw.com/. We checked that CLOTH does not contain sentence completion example questions from GRE, SAT and PSAT. }. 
Before cleaning, there are $20,605$ passages and $332,755$ questions. 
We perform the following processes to ensure the validity of data: Firstly, we remove questions with an inconsistent format such as questions with more than four options. Then we filter all questions whose validity relies on external information such as pictures or tables. Further, we find that half of the total passages are duplicates and we delete those passages. Lastly, on one of the websites, the answers are stored as images. We use two OCR software programs\footnote{tesseract: https://github.com/tesseract-ocr; ABBYY FineReader: https://www.abbyy.com/en-us/finereader/} to extract the answers from images. We discard the questions when results from the two software are different. After the cleaning process, we obtain a clean dataset of $7,131$ passages and $99,433$ questions. 

Since high school questions are more difficult than middle school questions, we divide the datasets into \namem and \nameh, which stand for the middle school part and the high school part. We split $11\%$ of the data for both the test set and the development set. The detailed statistics of the whole dataset and two subsets are presented in Table \ref{tab:training_size}. Note that the questions were created to test non-native speakers, hence the vocabulary size is not very large. 

\begin{table*}[ht]
\centering
\footnotesize
\begin{tabular}{l|ccc|ccc|ccc}
\toprule
\multirow{2}{*}{Dataset} & \multicolumn{3}{c|}{\namem} & \multicolumn{3}{c|}{\nameh} & \multicolumn{3}{c}{\name (Total)} \\ 

                 & Train & Dev  & Test & Train & Dev  & Test & Train & Dev   & Test      \\ \midrule 
                 \# passages            & 2,341  & 355  & 335  & 3,172  & 450  & 478  & 5,513  & 805   & 813    \\ 

\# questions           & 22,056 & 3,273 & 3,198 & 54,794 & 7,794 & 8,318 & 76,850 & 11,067 & 11,516  \\ 

Vocab. size         & & {15,096} & & & {32,212} & & & {37,235} & \\ 
\midrule

Avg. \# sentence            & & {16.26} & & & 18.92 & & & 17.79 & \\ 
Avg. \# words               &  & 242.88 &  & & 365.1 & & & 313.16 & \\ 
\bottomrule
\end{tabular}
\caption{The statistics of the training, development and test sets of \namem (middle school questions), \nameh (high school questions) and \name}
\label{tab:training_size} 
\end{table*}

\subsection{Question Type Analysis}
In order to evaluate students' mastery of a language, teachers usually design tests in a way that questions cover different aspects of a language. Specifically, they first identify words in the passage that can examine students' knowledge in vocabulary, logic, or grammar. Then, they replace the words with blanks and prepare three incorrect but nuanced candidate options to make the test non-trivial. A sample passage is presented in Table \ref{tab:sample}.
\begin{table}[th]
{\footnotesize
{\bf Passage:} 
Nancy had just got a job as a secretary in a company. Monday was the first day she went to work, so she was very \_1\_ and arrived early. She \_2\_ the door open and found nobody there. "I am the \_3\_ to arrive." She thought and came to her desk. She was surprised to find a bunch of \_4\_ on it. They were fresh. She \_5\_ them and they were sweet. She looked around for a \_6\_ to put them in. "Somebody has sent me flowers the very first day!" she thought \_7\_ . " But who could it be?" she began to \_8\_ . The day passed quickly and Nancy did everything with \_9\_ interest. For the following days of the \_10\_ , the first thing Nancy did was to change water for the followers and then set about her work. 

Then came another Monday. \_11\_ she came near her desk she was overjoyed to see a(n) \_12\_ bunch of flowers there. She quickly put them in the vase, \_13\_ the old ones. The same thing happened again the next Monday. Nancy began to think of ways to find out the \_14\_ . On Tuesday afternoon, she was sent to hand in a plan to the \_15\_ . She waited for his directives at his secretary's \_16\_ . She happened to see on the desk a half-opened notebook, which \_17\_ : "In order to keep the secretaries in high spirits, the company has decided that every Monday morning a bunch of fresh flowers should be put on each secretary’s desk." Later, she was told that their general manager was a business management psychologist.

\vspace{1ex}

{\bf Questions:}

\begin{tabular}{l@{\hskip 0.05in}l@{\hskip 0.05in}l@{\hskip 0.05in}l@{\hskip 0.05in}l}
1. & A. depressed&B. encouraged&\textbf{C. excited}&D. surprised \\
2. & A. turned&\textbf{B. pushed}&C. knocked&D. forced \\
3. & A. last&B. second&C. third&\textbf{D. first} \\
4. & A. keys&B. grapes&\textbf{C. flowers}&D. bananas \\
5. & \textbf{A. smelled}&B. ate&C. took&D. held \\
6. & \textbf{A. vase}&B. room&C. glass&D. bottle \\
7. & A. angrily&B. quietly&C. strangely&\textbf{D. happily }\\
8. & A. seek&\textbf{B. wonder}&C. work&D. ask \\
9. & A. low&B. little&\textbf{C. great}&D. general \\
10. & A. month&B. period&C. year&\textbf{D. week} \\
11. & A. Unless&\textbf{B. When}&C. Since&D. Before \\
12. & A. old&B. red&C. blue&\textbf{D. new} \\
13. & A. covering&B. demanding&\textbf{C. replacing}&D. forbidding \\
14. & \textbf{A. sender}&B. receiver&C. secretary&D. waiter \\
15. & A. assistant&B. colleague&C. employee&\textbf{D. manager} \\
16. & A. notebook&\textbf{B. desk}&C. office&D. house \\
17. &\textbf{A. said}&B. written&C. printed&D. signed \\
\end{tabular}

\begin{tabular}{lllll}
\end{tabular}
}

\caption{A Sample passage from our dataset. Bold faces highlight the correct answers. There is only one best answer among four candidates, although several candidates may seem correct.}
\label{tab:sample}
\end{table}

To understand the abilities of assessment on this dataset, we divide questions into several types and label the proportion of each type.
According to English teachers who regularly create cloze test questions for English exams in China, there are largely three types: grammar, vocabulary and reasoning. Grammar questions are easily differentiated from other two categories. However, the teachers themselves cannot specify a clear distinction between reasoning questions and vocabulary questions since all questions require comprehending the words within the context and conducting some level of reasoning by recognizing incomplete information or conceptual overlap. 

Hence, we divided the questions except grammar questions based on the difficulty level for a machine to answer the question, following works on analyzing machine comprehension datasets~\citep{chen2016thorough, trischler2016newsqa}. In particular, we divide them in terms of their dependency ranges, since questions that only involve a single sentence are easier to answer than questions involving evidence distributed in multiple sentences. Further, we divided questions involving long-term dependency into matching/paraphrasing questions and reasoning questions since matching questions are easier. The four types include:

\begin{itemize}[leftmargin=*]
\item Grammar: The question is about grammar usage, involving tense, preposition usage, active/passive voices, subjunctive mood and so on. 
\item Short-term-reasoning: The question is about content words and can be answered based on the information within the same sentence. Note that the content words can evaluate knowledge of both vocabulary and reasoning.
\item Matching/paraphrasing: The question is answered by copying/paraphrasing a word in the context. 
\item Long-term-reasoning: The answer must be inferred from synthesizing information distributed across multiple sentences. 
\end{itemize}

We sample $100$ passages in the high school category and the middle school category respectively with totally $3,000$ questions. 
The types of these questions are labeled on Amazon Turk. We pay \$1 and \$0.5 for high school passages and middle school passages respectively. We refer readers to Appendix \ref{sec:turker_label} for details of the labeling processes and the labeled sample passage.

The proportion of different questions is shown in Table \ref{tab:question_type}. 
The majority of questions are short-term-reasoning questions while approximately $22.4\%$ of the data needs long-term information, in which the long-term-reasoning questions constitute a large proportion.

\begin{table}[ht]
\footnotesize
\centering
\begin{tabular}{l|cc|cc|c}
\toprule
& \multicolumn{2}{|c|}{Short-term}& \multicolumn{2}{|c|}{Long-term}& \\ \midrule
Dataset & GM & STR & MP & LTR & O \\ \midrule
\name & 0.265 & 0.503 & 0.044 & 0.180 & 0.007 \\ 
\namem & 0.330&	0.413&	0.068&	0.174&	0.014 \\
\nameh & 0.240& 0.539	&0.035&	0.183&	0.004 \\
\bottomrule
\end{tabular}
\caption{The question type statistics of $3000$ sampled questions where GM, STR, MP, LTR and O denotes grammar, short-term-reasoning, matching/paraphrasing, long-term-reasoning and others respectively.   }
\label{tab:question_type}
\end{table}

\section{Exploring Models' Limits}
In this section, we investigate if human-created cloze test is a challenging problem for state-of-the-art models. We find that LM trained on the One Billion Word Corpus can achieve a remarkable score but cannot solve the cloze test. After conducting an error analysis, we hypothesize that the model is not able to deal with long-term dependencies. We verify the hypothesis by comparing the model's performance with the human performance when the information humans obtain is limited to one sentence. 

\subsection{Human and Model Performance}
\label{sec:human_model_performance}
\paragraph{LSTM} To test the performance of RNN-based supervised models, we train a bidirectional LSTM~\citep{hochreiter1997long} to predict the missing word given the context with only labeled data. The implementation details are in Appendix \ref{sec:implementation}. 
\paragraph{Attentive Readers} To enable the model to gather information from a longer context, we augment the supervised LSTM model with the attention mechanism~\citep{bahdanau2014neural}, so that the representation at the blank is used as a query to find the relevant context in the document and a blank-specific representation of the document is used to score each candidate answer. 
Specifically, we adapt the Stanford Attentive Reader~\citep{chen2016thorough} and the position-aware attention model~\citep{zhang2017position} to the cloze test problem. With the position-aware attention model, the attention scores are based on both the context match and the distance from a context to the blank. 
Both attention models are trained only with human-created blanks just as the LSTM model. 

\paragraph{LM} In cloze test, the context on both sides may be enough to determine the correct answer. Suppose $x_i$ is the missing word and $x_1, \cdots, x_{i-1}$, $x_{i+1}, \cdots, x_{n}$ are the context, 
we choose $x_i$ that maximizes the joint probability $p(x_1, \cdots, x_{n})$, which essentially maximizes the conditional likelihood $p(x_{i} \mid x_1, \cdots, x_{i-1}, x_{i+1}, \cdots, x_{n})$. Therefore, LM can be naturally adapted to cloze test. 

In essence, LM treats each word as a possible blank and learns to predict it. As a result, it receives more supervision than the LSTM trained on human-labeled questions. Besides training a neural LM on our dataset, interested in whether the state-of-the-art LM can solve cloze test, we also test the LM trained on the One Billion Word Benchmark~\citep{chelba2013one} (referred as 1B-LM) that achieves a perplexity of $30.0$~\citep{jozefowicz2016exploring}\footnote{The pre-trained model is obtained from https://github.com/tensorflow/models/tree/master/research/lm\_1b}. 
To make the evaluation time tractable, we limit the context length to one sentence or three sentences. Note that the One Billion Word Corpus does not overlap with the CLOTH corpus. 

\paragraph{Human performance} We measure the performance of Amazon Mechanical Turkers on $3,000$ sampled questions when the whole passage is given. 

\begin{table}[ht]
\centering
\footnotesize
\begin{tabular}{l|c@{\hskip 0.07in}c@{\hskip 0.07in}c}
\toprule
Model & \name & \namem & \nameh   \\ \midrule 
LSTM  & 0.484 & 0.518 & 0.471 \\
Stanford AR  & 0.487 & 0.529 & 0.471 \\
Position-aware AR  & 0.485 & 0.523 & 0.471 \\
\midrule
LM &0.548 & 0.646 & 0.506
\\
1B-LM (one sent.)  & 0.695 &0.723  & 0.685 \\
1B-LM (three sent.)  & 0.707 & {0.745} & {0.693} \\
\midrule
Human performance &0.859 & 0.897 & 0.845 \\
\bottomrule
\end{tabular}
\caption{Models' performance and human-level performance on \name. LSTM, Stanford Attentive Reader and Attentive Reader with position-aware attention shown in the top part only use supervised data labelled by human.  
LM outperforms LSTM since it receives more supervisions in learning to predict each word. Training on large external corpus further significantly enhances LM's accuracy.}
\label{tab:lm}
\end{table}

\paragraph{Results} The comparison is shown in Table ~\ref{tab:lm}. Both attentive readers achieve similar accuracy to the LSTM. We hypothesize that the reason of the attention model's unsatisfactory performance is that the evidence of a question cannot be simply found by matching the context. Similarly, on reading comprehension, though attention-based models~\citep{wang2017gated, seo2016bidirectional, dhingra2016gated} have reached human performance on the SQuAD dataset~\citep{rajpurkar2016squad}, their performance is still not comparable to human performance on datasets that focus more on reasoning where the evidence cannot be simply found by a matching behavior~\citep{lai2017large, xu2017towards}. Since the focus of this paper is to analyze the proposed dataset, we leave the design of reasoning oriented attention models for future work. 

The LM achieves much better performance than LSTM. The gap is larger when the LM is trained on the 1 Billion Word Corpus, indicating that more training data results in a better generalization. 
Specifically, the accuracy of 1B-LM is $0.695$ when one sentence is used as the context. It indicates that LM can learn sophisticated language regularities when given sufficient data. The same conclusion can also be drawn from the success of a concurrent work ELMo which uses LM representations as word vectors and achieves state-of-the-art results on six language tasks~\citep{peters2018deep}. 
However, if we increase the context length to three sentences, the accuracy of 1B-LM only has a marginal improvement. In contrast, humans outperform 1B-LM by a significant margin, which demonstrates that deliberately designed questions in \name are not completely solved even for state-of-the-art models.

\begin{table*}[th]

\vspace*{-3mm}
\scriptsize
	\centering
    \begin{tabular}{l|cccc} \toprule
    Context & \multicolumn{4}{c}{Options} \\ \midrule 
        She pushed the door open and found nobody there. "I am the \_\_ to arrive." She & \multirow{2}{*}{\textit{A. last}} & \multirow{2}{*}{B. second} & \multirow{2}{*}{C. third} & \multirow{2}{*}{\textbf{D. first}} \\ 
    thought and came to her desk. \\ \midrule
    They were fresh. She \_\_ them and they were sweet. She looked around for a vase &\multirow{2}{*}{\textbf{A. smelled}}&\multirow{2}{*}{\textit{B. ate}} &\multirow{2}{*}{C. took}&\multirow{2}{*}{D. held} \\
     to put them in. \\
    \midrule
    She smelled them and they were sweet. She looked around for a \_\_ to put them in. & \multirow{2}{*}{\textbf{A. vase}}&\multirow{2}{*}{\textit{B. room}}&\multirow{2}{*}{C. glass}&\multirow{2}{*}{D. bottle} \\ 
    "Somebody has sent me flowers the very first day!"\\    \midrule
    "But who could it be?" she began to \_\_ . The day passed quickly and Nancy did  &\multirow{2}{*}{A. seek}&\multirow{2}{*}{\textbf{B. wonder}}&\multirow{2}{*}{C. work}& \multirow{2}{*}{\textit{D. ask}} \\
    everything with great interest.  \\ \bottomrule
    \end{tabular}
    \caption{Error analysis of 1-billion-language-model with three sentences as the context. The questions are sampled from the sample passage shown in Table \ref{tab:sample}. The correct answer is in bold text. The incorrectly selected options are in italics. }
    \label{tab:error}
\end{table*}
\subsection{Analyzing 1B-LM's Strengths and Weaknesses}
\label{sec:analyze}
In this section, we would like to understand why 1B-LM lags behind human performance. 
We find that most of the errors involve long-term reasoning. Additionally, in a lot of cases, the dependency is within the context of three sentences. 
We show several errors made by the 1B-LM in Table \ref{tab:error}.
In the first example, the model does not know that Nancy found nobody in the company means that Nancy was the first one to arrive at the company. 
In the second and third example, the model fails probably because of not recognizing ``they" referred to ``flowers". The dependency in the last case is longer. It depends on the fact that Nancy was alone in the company.

Based on the case study, we hypothesize that the LM is not able to take long-term information into account, although it achieves a surprisingly good overall performance. Additionally, the 1B-LM is trained on the sentence level, which might also result in the inability to track paragraph level information.
However, to investigate the differences between training on sentence level and on paragraph level, a prohibitive amount of computational resource is required to train a large model on the 1 Billion Word Corpus. 

On the other hand, a practical comparison is to test the model's performance on different types of questions. We find that the model's accuracy is $0.591$ on long-term-reasoning questions of \nameh while it achieves $0.693$ on short-term-reasoning (a comprehensive type-specific performance is available in Appendix \ref{sec:implementation}), which partially confirms that long-term-reasoning is harder. However, we could not completely rely on the performance on specific questions types, partly due to a large variance caused by the small sample size. Another reason is that the reliability of question type labels depends on whether turkers are careful enough. For example, in the error analysis shown in Table \ref{tab:error}, a careless turker would label the second example as short-term-reasoning without noticing that the meaning of ``they" relies on a long context. 

To objectively verify if the LM's strengths lie in dealing with short-term information, we obtain the ceiling performance of only utilizing short-term information. Showing only one sentence as the context, 
we ask the Turkers to select an option based on their best guesses given the insufficient information. By limiting the context span manually, the ceiling performance with  the access to only a short context is estimated accurately. 

\begin{table}[ht]
\centering
\footnotesize
\begin{tabular}{l|l|c@{\hskip 0.04in}c@{\hskip 0.04in}c}
\toprule
&Model & \name & \namem & \nameh   \\ \midrule 
\multirow{2}{*}{Short context}& 1B-LM & 0.695 &0.723  & 0.685 \\
&Human & 0.713 & 0.771 & 0.691  \\
\midrule
\multirow{2}{*}{Long context} & 1B-LM  & 0.707 & {0.745} & {0.693} \\
&Human &0.859 & 0.897 & 0.845 \\
\bottomrule
\end{tabular}

\caption{Humans' performance compared with 1-billion-language-model. In the short context part, both 1B-LM and humans only use information of one sentence. In the long context part, humans have the whole passage as the context, while 1B-LM uses contexts of three sentences.}
\label{tab:human_performance}
\end{table}

As shown in Table \ref{tab:human_performance}, The performance of 1B-LM using one sentence as the context can almost match the human ceiling performance of only using short-term information. Hence we conclude that the LM can almost perfectly solve all short-term cloze questions. However, the performance of LM is not improved significantly when a long-term context is given, indicating that the performance gap is due to the inability of long-term reasoning.

\section{Comparing Human-created Data and Automatically-generated Data}
\label{sec:difficulty}
In this section, we demonstrate that human-created data is a better testbed than automatically-generated cloze test since it results in a larger gap between model's performance and human performance. 

A casual observation is that a cloze test can be created by randomly deleting words and randomly sampling candidate options. In fact, to generate large-scale data, similar generation processes have been introduced and widely used in machine comprehension ~\citep{hermann2015teaching, hill2015goldilocks, onishi2016did}. However, research on cloze test design~\citep{sachs1997construct} shows that tests created by deliberately deleting words are more reliable than tests created by randomly or periodically deleting words. To design accurate language proficiency assessment, teachers usually deliberately select words in order to examine students' proficiency in grammar, vocabulary and reasoning. Moreover, in order to make the question non-trivial, three incorrect options provided by teachers are usually grammatically correct and relevant to the context. For instance, in the fourth problem of the sample passage shown in Table \ref{tab:sample}, ``grapes'', ``flowers'' and ``bananas'' all fit the description of being fresh. 

Hence we naturally hypothesize that human-generated data has distinct characteristics when compared with automatically-generated data. To verify this assumption, we compare the LSTM model's performance when given different proportions of the two types of data.
Specifically, to train a model with $\alpha$ percent of automatically-generated data, we randomly replace $a$ percent blanks with blanks at random positions, while keeping the remaining $1-\alpha$ percent questions the same. The candidate options for the generated blanks are random words sampled from the unigram distribution. We test models obtained with varying $\alpha$ on human-created data and automatically-generated data respectively.  

\begin{table}[ht]
\centering
\footnotesize
\vspace*{-3mm}
\begin{tabular}{c|c@{\hskip 0.08in}c@{\hskip 0.08in}c@{\hskip 0.08in}c@{\hskip 0.08in}c}
\toprule
\backslashbox{Test}{$\alpha \%$}  & $0\%$ &  $25\%$ & $50\%$ & $75\%$ & $100\%$ \\
\midrule
human-created & 0.484 & 0.475 & 0.469 & 0.423  & 0.381 \\
Generated & 0.422 & 0.699 & 0.757 & 0.785 & 0.815 \\
\bottomrule
\end{tabular}

\caption{The model's performance when trained on $\alpha$ percent of automatically-generated data and $100-\alpha$ percent of human-created data}
\label{tab:diff_new}

\end{table}
From the comparison in Table \ref{tab:diff_new}, we have the following observations: (1) human-created data leads to a larger gap between model's performance and the ceiling/human performance. The model's performance and human's performance on the human-created data are $0.484$ and $0.859$ respectively, as shown in Tab. \ref{tab:lm}, leading to a gap of $0.376$. In comparison, the performance gap on the automatically-generated data is at most $0.185$ since the model's performance reaches an accuracy of $0.815$ when fully trained on generated data. (2) Although human-created data may provide more information in distinguishing similar words, the distributional mismatch between two types of data makes it non-trivial to transfer the knowledge gained from human-created data to tackle automatically-generated data. Specifically, the model's performance on automatically-generated data monotonically decreases when given a higher ratio of human-created data. 


\section{Combining Human-created Data with Automatically-generated Data}
\label{sec:combine} 

In Section \ref{sec:human_model_performance}, we show that LM is able to take advantage of more supervision since it predicts each word based on the context. At the same time, we also show that human-created data and the automatically-generated data are quite different in Section \ref{sec:difficulty}. 
In this section, we propose a model that takes advantage of both sources. 

\subsection{Representative-based Model} 
\label{sec:rep_model}
Specifically, for each question, regardless of being human-created or automatically-generated, we can compute the negative log likelihood of the correct answer as the loss function.
Suppose $J_\mathrm{H}$ is the average negative log likelihood loss for human-created questions and $J_\mathrm{R}$ is the loss function on generated questions, we combine losses on human-created questions and generated questions by simply adding them together, i.e., $J_\mathrm{R}+J_\mathrm{H}$ is used as the final loss function. We will introduce the definition of $J_\mathrm{R}$ in the following paragraphs.

Although automatically-generated data has a large quantity and is valuable to the model training, as shown in the previous Section, automatically-generated questions are quite different from human-created questions. Ideally, a large amount of human-created questions is more desirable than a large amount of automatically-generated questions. A possible avenue towards having large-scale human-created data is to automatically pick out a large number of generated questions which are \textit{representative} of or similar to human-created questions. 
In other words, we train a network to predict whether a question is a generated question or a human-created question. A generated question is representative of human-created questions if it has a high probability of being a human-created question. Then we can give higher weights to questions that resemble human-created question. 

We first introduce our method to obtain the representativeness information. Let $x$ denote the passage and $z$ denote whether a word is selected as a question by human, i.e., $z$ is $1$ if this word is selected to be filled in the original passage or $0$ otherwise. Suppose $h_i$ is the representation of $i$-th word given by a bidirectional LSTM. The network computes the probability $p_i$ of $x_i$ being a human-created question as follows:

\begin{equation*}
l_i = h^T_i w_{x_i}; \quad p_i = \mbox{Sigmoid}(l_i) 
\end{equation*}
where $l_i$ is the logit which will be used as in the final model and $w_{x_i}$ is the the word embedding.
We train the network to minimize the binary cross entropy between $p$ and ground-truth labels at each token. 

After obtaining the representativeness information, we define the representativeness weighted loss function as
$$
J_\mathrm{R} = \sum_{i \not\in H} \mbox{Softmax}_i(\frac{l_1}{\alpha}, \cdots, \frac{l_n}{\alpha})J_{i}
$$
where $J_i$ denotes the negative log likelihood loss for the $i-$th question and let $l_i$ be the output representativeness of the $i$-th question and $H$ is the set of all human-generated questions and $\alpha$ is the temperature of the Softmax function. The model degenerates into assigning a uniform weight to all questions when the temperature is $+\infty$. We set $\alpha$ to $2$ based on the performance on the dev set. \footnote{The code is available at https://github.com/qizhex/Large-scale-Cloze-Test-Dataset-Created-by-Teachers}.

\begin{figure*}[h!]
\centering
\includegraphics[scale=0.25]{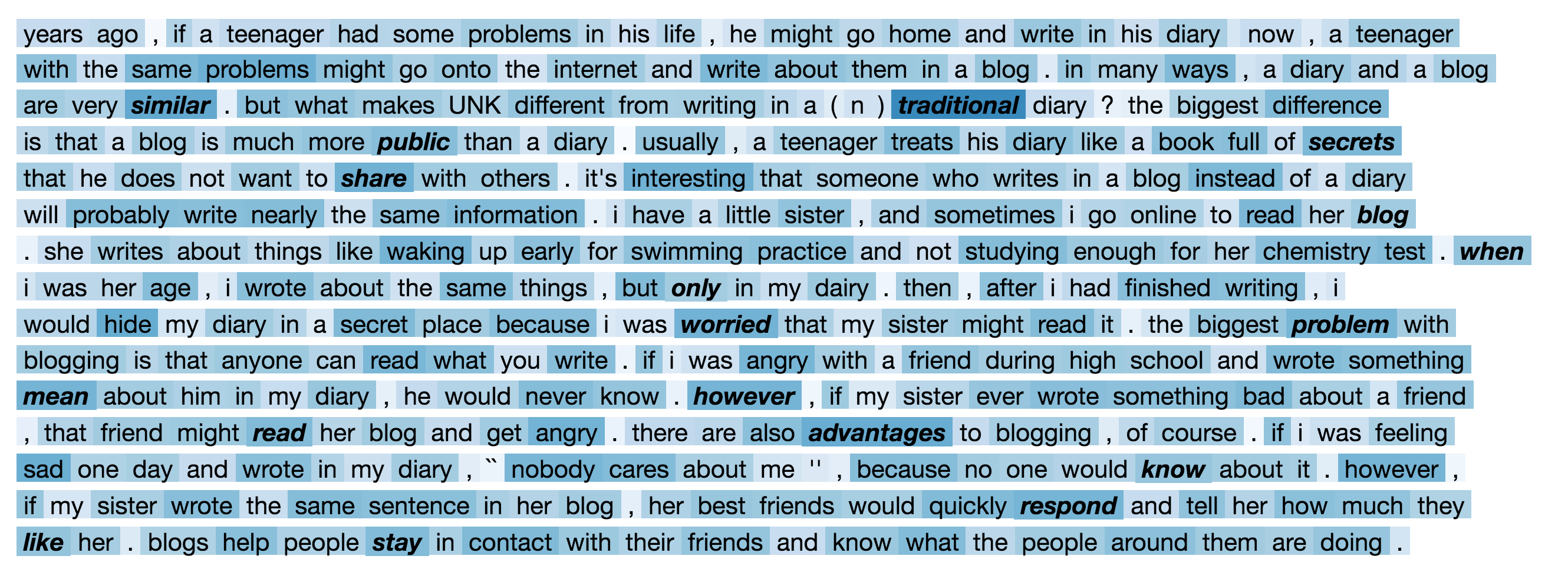}
\caption{Representativeness prediction for each word. Lighter color means less representative. The words deleted by human as blanks are in bold text.  }
\label{fig:generated}
\end{figure*}

\begin{table}[ht]
\centering
\footnotesize
\begin{tabular}{l|c|ccc}
\toprule
Model & Ex. & \name & \namem & \nameh   \\
\midrule
Our model & \multirow{4}{*}{No} & \textbf{0.583} & \textbf{0.673} & \textbf{0.549} \\
LM & &0.548 & 0.646 & 0.506 \\
LSTM & &  0.484 & 0.518 & 0.471 \\
Stanford AR &  & 0.487 & 0.529 & 0.471 \\
\midrule
 1B-LM & Yes  & 0.707 & 0.745 & 0.693 \\
\midrule
 Human &  & 0.859 & 0.897 & 0.845   \\ 
\bottomrule
\end{tabular}
\caption{Overall results on \name. Ex. denotes external data. }
\label{tab:main_result}
\end{table}

\begin{table}[ht]
\centering
\vspace*{-3mm}
\footnotesize
\begin{tabular}{l|c@{\hskip 0.08in}c@{\hskip 0.08in}c}
\toprule
Model & \name & \namem & \nameh   \\
\midrule
Our model & \textbf{0.583} & \textbf{0.673} & \textbf{0.549} \\
\quad w.o. rep.  & 0.566  & 0.662 & 0.528 \\
\quad w.o. hum. & 0.565 & 0.665 & 0.526\\
\quad w.o. rep. or hum.  & 0.543 & 0.643 & 0.505   \\
\bottomrule
\end{tabular}
\caption{Ablation study on using the representativeness information (denoted as rep.) and the human-created data (denoted as hum.)}
\label{tab:ablation}
\end{table}

\subsection{Results} We summarize performances of all models in Table \ref{tab:main_result}. Our representativeness model outperforms all other models that do not use external data on \name, \nameh and \namem. 

\subsection{Analysis}
In this section, we verify the effectiveness of the representativeness-based averaging by ablation studies. When we remove the representativeness information by setting $\alpha$ to infinity, the accuracy drops from $0.583$ to $0.566$. When we further remove the human-created data so that only generated data is employed, the accuracy drops to $0.543$, similar to the performance of LM. The results further confirm that it is beneficial to incorporate human-created questions into training. 

A sample of the predicted representativeness is shown in Figure \ref{fig:generated}\footnote{The script to generate the Figure is obtained at \url{https://gist.github.com/ihsgnef/f13c35cd46624c8f458a4d23589ac768}}.  
Clearly, words that are too obvious have low scores, such as punctuation marks, simple words ``a" and ``the". In contrast, content words whose semantics are directly related to the context have a higher score, e.g., ``same", ``similar", ``difference" have a high score when the difference between two objects is discussed and ``secrets" has a high score since it is related to the subsequent sentence ``does not want to share with others". 
Our prediction model achieves an F1 score of $36.5$ on the test set, which is understandable since there are many plausible questions within a passage.

It has been shown that features such as morphology information and readability are beneficial in cloze test prediction~\citep{skory2010predicting, correia2012automatic, correia2010automatic, kurtasov2013system}. We leave investigating the advanced approaches of automatically designing cloze test to future work.

\section{Conclusion and Discussion}
In this paper, we propose a large-scale cloze test dataset \name that is designed by teachers. 
With missing blanks and candidate options carefully created by teachers to test different aspects of language phenomena, \name requires a deep language understanding and better captures the complexity of human language. 
We find that human outperforms 1B-LM by a significant margin. After detailed analysis, we find that the performance gap is due to the model's inability to understanding a long context. We also show that, compared to automatically-generated questions, human-created questions are more difficult and lead to a larger margin between human performance and the model's performance. 

Despite the excellent performance of 1B-LM when compared with models trained  only on \name, it is still important to investigate and create more effective models and algorithms which provide complementary advantages to having a large amount of data. For rapid algorithm developments, we suggest training models only on the training set of \name and comparing with models that do not utilize external data. 

We hope our dataset provides a valuable testbed to the language modeling community and the machine comprehension community. 
In particular, the language modeling community can use CLOTH to evaluate their models' abilities in modeling a long context. In addition, the machine comprehension community may also find CLOTH useful in evaluating machine's understanding of language phenomena including vocabulary, reasoning and grammar, which are key components of comprehending natural language.

In our future work, we would like to design algorithms to better model a long context, to utilize external knowledge, and to explore more effective semi-supervised learning approaches.
Firstly,  we would like to investigate efficient ways of utilizing external knowledge such as paraphrasing and semantic concepts like prior works~\citep{dong2017learning, dasigi2017ontology}. In comparison, training on a large external dataset is actually a time-consuming way of utilizing external knowledge.
Secondly, to use the generated questions more effectively, the representative-based semi-supervised approach might be improved by techniques studied in active learning and hard example mining~\citep{Settles2009ActiveLL, shrivastava2016training, chang2017active}. 

\section*{Acknowledgement}
We thank Yulun Du, Kaiyu Shi and Zhilin Yang for insightful discussions and suggestions on the draft. We thank Shi Feng for the script to highlight representative words. This research was supported in part by DARPA grant FA8750-12-2-0342 funded under the DEFT program.

\bibliography{bib}

\begin{thebibliography}{44}
\expandafter\ifx\csname natexlab\endcsname\relax\def\natexlab#1{#1}\fi

\bibitem[{Bahdanau et~al.(2014)Bahdanau, Cho, and Bengio}]{bahdanau2014neural}
Dzmitry Bahdanau, Kyunghyun Cho, and Yoshua Bengio. 2014.
\newblock Neural machine translation by jointly learning to align and
  translate.
\newblock \emph{arXiv preprint arXiv:1409.0473}.

\bibitem[{Chang et~al.(2017)Chang, Learned-Miller, and
  McCallum}]{chang2017active}
Haw-Shiuan Chang, Erik Learned-Miller, and Andrew McCallum. 2017.
\newblock Active bias: Training more accurate neural networks by emphasizing
  high variance samples.
\newblock In \emph{Advances in Neural Information Processing Systems}, pages
  1003--1013.

\bibitem[{Chelba et~al.(2013)Chelba, Mikolov, Schuster, Ge, Brants, Koehn, and
  Robinson}]{chelba2013one}
Ciprian Chelba, Tomas Mikolov, Mike Schuster, Qi~Ge, Thorsten Brants, Phillipp
  Koehn, and Tony Robinson. 2013.
\newblock One billion word benchmark for measuring progress in statistical
  lming.
\newblock \emph{arXiv preprint arXiv:1312.3005}.

\bibitem[{Chen et~al.(2016)Chen, Bolton, and Manning}]{chen2016thorough}
Danqi Chen, Jason Bolton, and Christopher~D Manning. 2016.
\newblock A thorough examination of the cnn/daily mail reading comprehension
  task.
\newblock \emph{arXiv preprint arXiv:1606.02858}.

\bibitem[{Clark et~al.(2018)Clark, Cowhey, Etzioni, Khot, Sabharwal, Schoenick,
  and Tafjord}]{clark2018think}
Peter Clark, Isaac Cowhey, Oren Etzioni, Tushar Khot, Ashish Sabharwal, Carissa
  Schoenick, and Oyvind Tafjord. 2018.
\newblock Think you have solved question answering? try arc, the ai2 reasoning
  challenge.
\newblock \emph{arXiv preprint arXiv:1803.05457}.

\bibitem[{Correia et~al.(2012)Correia, Baptista, Eskenazi, and
  Mamede}]{correia2012automatic}
Rui Correia, Jorge Baptista, Maxine Eskenazi, and Nuno~J Mamede. 2012.
\newblock Automatic generation of cloze question stems.
\newblock In \emph{PROPOR}, pages 168--178. Springer.

\bibitem[{Correia et~al.(2010)Correia, Baptista, Mamede, Trancoso, and
  Eskenazi}]{correia2010automatic}
Rui Correia, Jorge Baptista, Nuno Mamede, Isabel Trancoso, and Maxine Eskenazi.
  2010.
\newblock Automatic generation of cloze question distractors.
\newblock In \emph{Proceedings of the Interspeech 2010 Satellite Workshop on
  Second Language Studies: Acquisition, Learning, Education and Technology,
  Waseda University, Tokyo, Japan}.

\bibitem[{Dasigi et~al.(2017)Dasigi, Ammar, Dyer, and
  Hovy}]{dasigi2017ontology}
Pradeep Dasigi, Waleed Ammar, Chris Dyer, and Eduard Hovy. 2017.
\newblock Ontology-aware token embeddings for prepositional phrase attachment.
\newblock \emph{arXiv preprint arXiv:1705.02925}.

\bibitem[{Dhingra et~al.(2016)Dhingra, Liu, Yang, Cohen, and
  Salakhutdinov}]{dhingra2016gated}
Bhuwan Dhingra, Hanxiao Liu, Zhilin Yang, William~W Cohen, and Ruslan
  Salakhutdinov. 2016.
\newblock Gated-attention readers for text comprehension.
\newblock \emph{arXiv preprint arXiv:1606.01549}.

\bibitem[{Dong et~al.(2017)Dong, Mallinson, Reddy, and
  Lapata}]{dong2017learning}
Li~Dong, Jonathan Mallinson, Siva Reddy, and Mirella Lapata. 2017.
\newblock Learning to paraphrase for question answering.
\newblock \emph{arXiv preprint arXiv:1708.06022}.

\bibitem[{Fotos(1991)}]{fotos1991cloze}
Sandra~S Fotos. 1991.
\newblock The cloze test as an integrative measure of efl proficiency: A
  substitute for essays on college entrance examinations?
\newblock \emph{Language Learning}, 41(3):313--336.

\bibitem[{Hermann et~al.(2015)Hermann, Kocisky, Grefenstette, Espeholt, Kay,
  Suleyman, and Blunsom}]{hermann2015teaching}
Karl~Moritz Hermann, Tomas Kocisky, Edward Grefenstette, Lasse Espeholt, Will
  Kay, Mustafa Suleyman, and Phil Blunsom. 2015.
\newblock Teaching machines to read and comprehend.
\newblock In \emph{NIPS}.

\bibitem[{Hill et~al.(2016)Hill, Bordes, Chopra, and
  Weston}]{hill2015goldilocks}
Felix Hill, Antoine Bordes, Sumit Chopra, and Jason Weston. 2016.
\newblock The goldilocks principle: Reading children's books with explicit
  memory representations.
\newblock \emph{ICLR}.

\bibitem[{Hochreiter and Schmidhuber(1997)}]{hochreiter1997long}
Sepp Hochreiter and J{\"u}rgen Schmidhuber. 1997.
\newblock Long short-term memory.
\newblock \emph{Neural computation}, 9(8):1735--1780.

\bibitem[{Jonz(1991)}]{jonz1991cloze}
Jon Jonz. 1991.
\newblock Cloze item types and second language comprehension.
\newblock \emph{Language testing}, 8(1):1--22.

\bibitem[{Joshi et~al.(2017)Joshi, Choi, Weld, and
  Zettlemoyer}]{joshi2017triviaqa}
Mandar Joshi, Eunsol Choi, Daniel~S Weld, and Luke Zettlemoyer. 2017.
\newblock Triviaqa: A large scale distantly supervised challenge dataset for
  reading comprehension.
\newblock \emph{ACL}.

\bibitem[{Jozefowicz et~al.(2016)Jozefowicz, Vinyals, Schuster, Shazeer, and
  Wu}]{jozefowicz2016exploring}
Rafal Jozefowicz, Oriol Vinyals, Mike Schuster, Noam Shazeer, and Yonghui Wu.
  2016.
\newblock Exploring the limits of lming.
\newblock \emph{arXiv preprint arXiv:1602.02410}.

\bibitem[{Kingma and Ba(2014)}]{kingma2014adam}
Diederik Kingma and Jimmy Ba. 2014.
\newblock Adam: A method for stochastic optimization.
\newblock \emph{arXiv preprint arXiv:1412.6980}.

\bibitem[{Kurtasov(2013)}]{kurtasov2013system}
Andrey Kurtasov. 2013.
\newblock A system for generating cloze test items from russian-language text.
\newblock In \emph{Proceedings of the Student Research Workshop associated with
  RANLP 2013}, pages 107--112.

\bibitem[{Lai et~al.(2017)Lai, Xie, Liu, Yang, and Hovy}]{lai2017large}
Guokun Lai, Qizhe Xie, Hanxiao Liu, Yiming Yang, and Eduard Hovy. 2017.
\newblock Race: Large-scale reading comprehension dataset from examinations.
\newblock \emph{EMNLP}.

\bibitem[{Nguyen et~al.(2016)Nguyen, Rosenberg, Song, Gao, Tiwary, Majumder,
  and Deng}]{nguyen2016ms}
Tri Nguyen, Mir Rosenberg, Xia Song, Jianfeng Gao, Saurabh Tiwary, Rangan
  Majumder, and Li~Deng. 2016.
\newblock Ms marco: A human generated machine reading comprehension dataset.
\newblock \emph{arXiv preprint arXiv:1611.09268}.

\bibitem[{Onishi et~al.(2016)Onishi, Wang, Bansal, Gimpel, and
  McAllester}]{onishi2016did}
Takeshi Onishi, Hai Wang, Mohit Bansal, Kevin Gimpel, and David McAllester.
  2016.
\newblock Who did what: A large-scale person-centered cloze dataset.
\newblock \emph{arXiv preprint arXiv:1608.05457}.

\bibitem[{Paperno et~al.(2016)Paperno, Kruszewski, Lazaridou, Pham, Bernardi,
  Pezzelle, Baroni, Boleda, and Fern{\'a}ndez}]{paperno2016lambada}
Denis Paperno, Germ{\'a}n Kruszewski, Angeliki Lazaridou, Quan~Ngoc Pham,
  Raffaella Bernardi, Sandro Pezzelle, Marco Baroni, Gemma Boleda, and Raquel
  Fern{\'a}ndez. 2016.
\newblock The lambada dataset: Word prediction requiring a broad discourse
  context.
\newblock \emph{arXiv preprint arXiv:1606.06031}.

\bibitem[{Paszke et~al.(2017)Paszke, Gross, Chintala, Chanan, Yang, DeVito,
  Lin, Desmaison, Antiga, and Lerer}]{paszke2017automatic}
Adam Paszke, Sam Gross, Soumith Chintala, Gregory Chanan, Edward Yang, Zachary
  DeVito, Zeming Lin, Alban Desmaison, Luca Antiga, and Adam Lerer. 2017.
\newblock Automatic differentiation in pytorch.
\newblock In \emph{NIPS-W}.

\bibitem[{Pe{\~n}as et~al.(2014)Pe{\~n}as, Miyao, Rodrigo, Hovy, and
  Kando}]{penas2014overview}
Anselmo Pe{\~n}as, Yusuke Miyao, {\'A}lvaro Rodrigo, Eduard~H Hovy, and Noriko
  Kando. 2014.
\newblock Overview of clef qa entrance exams task 2014.
\newblock In \emph{CLEF (Working Notes)}, pages 1194--1200.

\bibitem[{Pennington et~al.(2014)Pennington, Socher, and
  Manning}]{pennington2014glove}
Jeffrey Pennington, Richard Socher, and Christopher Manning. 2014.
\newblock Glove: Global vectors for word representation.
\newblock In \emph{EMNLP}, pages 1532--1543.

\bibitem[{Peters et~al.(2018)Peters, Neumann, Iyyer, Gardner, Clark, Lee, and
  Zettlemoyer}]{peters2018deep}
Matthew~E Peters, Mark Neumann, Mohit Iyyer, Matt Gardner, Christopher Clark,
  Kenton Lee, and Luke Zettlemoyer. 2018.
\newblock Deep contextualized word representations.
\newblock \emph{arXiv preprint arXiv:1802.05365}.

\bibitem[{Rajpurkar et~al.(2016)Rajpurkar, Zhang, Lopyrev, and
  Liang}]{rajpurkar2016squad}
Pranav Rajpurkar, Jian Zhang, Konstantin Lopyrev, and Percy Liang. 2016.
\newblock Squad: 100,000+ questions for machine comprehension of text.
\newblock \emph{arXiv preprint arXiv:1606.05250}.

\bibitem[{Rodrigo et~al.(2015)Rodrigo, Pe{\~n}as, Miyao, Hovy, and
  Kando}]{rodrigo2015overview}
{\'A}lvaro Rodrigo, Anselmo Pe{\~n}as, Yusuke Miyao, Eduard~H Hovy, and Noriko
  Kando. 2015.
\newblock Overview of clef qa entrance exams task 2015.
\newblock In \emph{CLEF (Working Notes)}.

\bibitem[{Sachs et~al.(1997)Sachs, Tung, and Lam}]{sachs1997construct}
J~Sachs, P~Tung, and RYH Lam. 1997.
\newblock How to construct a cloze test: Lessons from testing measurement
  theory models.
\newblock \emph{Perspectives}.

\bibitem[{Schoenick et~al.(2017)Schoenick, Clark, Tafjord, Turney, and
  Etzioni}]{schoenick2017moving}
Carissa Schoenick, Peter Clark, Oyvind Tafjord, Peter Turney, and Oren Etzioni.
  2017.
\newblock Moving beyond the turing test with the allen ai science challenge.
\newblock \emph{Communications of the ACM}, 60(9):60--64.

\bibitem[{Seo et~al.(2016)Seo, Kembhavi, Farhadi, and
  Hajishirzi}]{seo2016bidirectional}
Minjoon Seo, Aniruddha Kembhavi, Ali Farhadi, and Hannaneh Hajishirzi. 2016.
\newblock Bidirectional attention flow for machine comprehension.
\newblock \emph{arXiv preprint arXiv:1611.01603}.

\bibitem[{Settles(2009)}]{Settles2009ActiveLL}
Burr Settles. 2009.
\newblock Active learning literature survey.

\bibitem[{Shibuki et~al.(2014)Shibuki, Sakamoto, Kano, Mitamura, Ishioroshi,
  Itakura, Wang, Mori, and Kando}]{shibuki2014overview}
Hideyuki Shibuki, Kotaro Sakamoto, Yoshinobu Kano, Teruko Mitamura, Madoka
  Ishioroshi, Kelly~Y Itakura, Di~Wang, Tatsunori Mori, and Noriko Kando. 2014.
\newblock Overview of the ntcir-11 qa-lab task.
\newblock In \emph{NTCIR}.

\bibitem[{Shrivastava et~al.(2016)Shrivastava, Gupta, and
  Girshick}]{shrivastava2016training}
Abhinav Shrivastava, Abhinav Gupta, and Ross Girshick. 2016.
\newblock Training region-based object detectors with online hard example
  mining.
\newblock In \emph{Proceedings of the IEEE Conference on Computer Vision and
  Pattern Recognition}, pages 761--769.

\bibitem[{Skory and Eskenazi(2010)}]{skory2010predicting}
Adam Skory and Maxine Eskenazi. 2010.
\newblock Predicting cloze task quality for vocabulary training.
\newblock In \emph{Proceedings of the NAACL HLT 2010 Fifth Workshop on
  Innovative Use of NLP for Building Educational Applications}, pages 49--56.
  Association for Computational Linguistics.

\bibitem[{Taylor(1953)}]{taylor1953cloze}
Wilson~L Taylor. 1953.
\newblock “cloze procedure”: a new tool for measuring readability.
\newblock \emph{Journalism Bulletin}, 30(4):415--433.

\bibitem[{Tremblay(2011)}]{tremblay2011proficiency}
Annie Tremblay. 2011.
\newblock Proficiency assessment standards in second language acquisition
  research.
\newblock \emph{Studies in Second Language Acquisition}, 33(3):339--372.

\bibitem[{Trischler et~al.(2016)Trischler, Wang, Yuan, Harris, Sordoni,
  Bachman, and Suleman}]{trischler2016newsqa}
Adam Trischler, Tong Wang, Xingdi Yuan, Justin Harris, Alessandro Sordoni,
  Philip Bachman, and Kaheer Suleman. 2016.
\newblock Newsqa: A machine comprehension dataset.
\newblock \emph{arXiv preprint arXiv:1611.09830}.

\bibitem[{Wang et~al.(2017)Wang, Yang, Wei, Chang, and Zhou}]{wang2017gated}
Wenhui Wang, Nan Yang, Furu Wei, Baobao Chang, and Ming Zhou. 2017.
\newblock Gated self-matching networks for reading comprehension and question
  answering.
\newblock In \emph{Proceedings of the 55th Annual Meeting of the Association
  for Computational Linguistics (Volume 1: Long Papers)}, volume~1, pages
  189--198.

\bibitem[{Xu et~al.(2017)Xu, Liu, Gao, Shen, and Liu}]{xu2017towards}
Yichong Xu, Jingjing Liu, Jianfeng Gao, Yelong Shen, and Xiaodong Liu. 2017.
\newblock Towards human-level machine reading comprehension: Reasoning and
  inference with multiple strategies.
\newblock \emph{arXiv preprint arXiv:1711.04964}.

\bibitem[{Zhang et~al.(2017)Zhang, Zhong, Chen, Angeli, and
  Manning}]{zhang2017position}
Yuhao Zhang, Victor Zhong, Danqi Chen, Gabor Angeli, and Christopher~D Manning.
  2017.
\newblock Position-aware attention and supervised data improve slot filling.
\newblock In \emph{Proceedings of the 2017 Conference on Empirical Methods in
  Natural Language Processing}, pages 35--45.

\bibitem[{Zhu et~al.(2015)Zhu, Kiros, Zemel, Salakhutdinov, Urtasun, Torralba,
  and Fidler}]{zhu2015aligning}
Yukun Zhu, Ryan Kiros, Rich Zemel, Ruslan Salakhutdinov, Raquel Urtasun,
  Antonio Torralba, and Sanja Fidler. 2015.
\newblock Aligning books and movies: Towards story-like visual explanations by
  watching movies and reading books.
\newblock In \emph{Proceedings of the IEEE international conference on computer
  vision}, pages 19--27.

\bibitem[{Zweig and Burges(2011)}]{zweig2011microsoft}
Geoffrey Zweig and Christopher~JC Burges. 2011.
\newblock The microsoft research sentence completion challenge.
\newblock Technical report, Technical Report MSR-TR-2011-129, Microsoft.

\end{thebibliography}
\bibliographystyle{acl_natbib}

\clearpage
\appendix
\onecolumn
\clearpage
\section{Appendix}
\subsection{Question Type Labeling}
\label{sec:turker_label}
To label the questions, we provided the definition and an example for each question category to the Amazon Mechanical Turkers. To ensure quality, we limited the workers to master Turkers who are experienced and maintain a high acceptance rate. However, we did not restrict the backgrounds of the Turkers since master Turkers should have a reasonable amount of knowledge about English to conduct previous tasks. In addition, the vocabulary used in \name are usually not difficult since they are constructed to test non-native speakers in middle school or high school. 
To get a concrete idea of the nature of question types, please refer to examples shown in Tab. \ref{tab:sample_label}.

\begin{figure*}[t!]
    \centering
    \begin{subfigure}[t]{0.5\textwidth}
        \centering
        \includegraphics[scale=0.4]{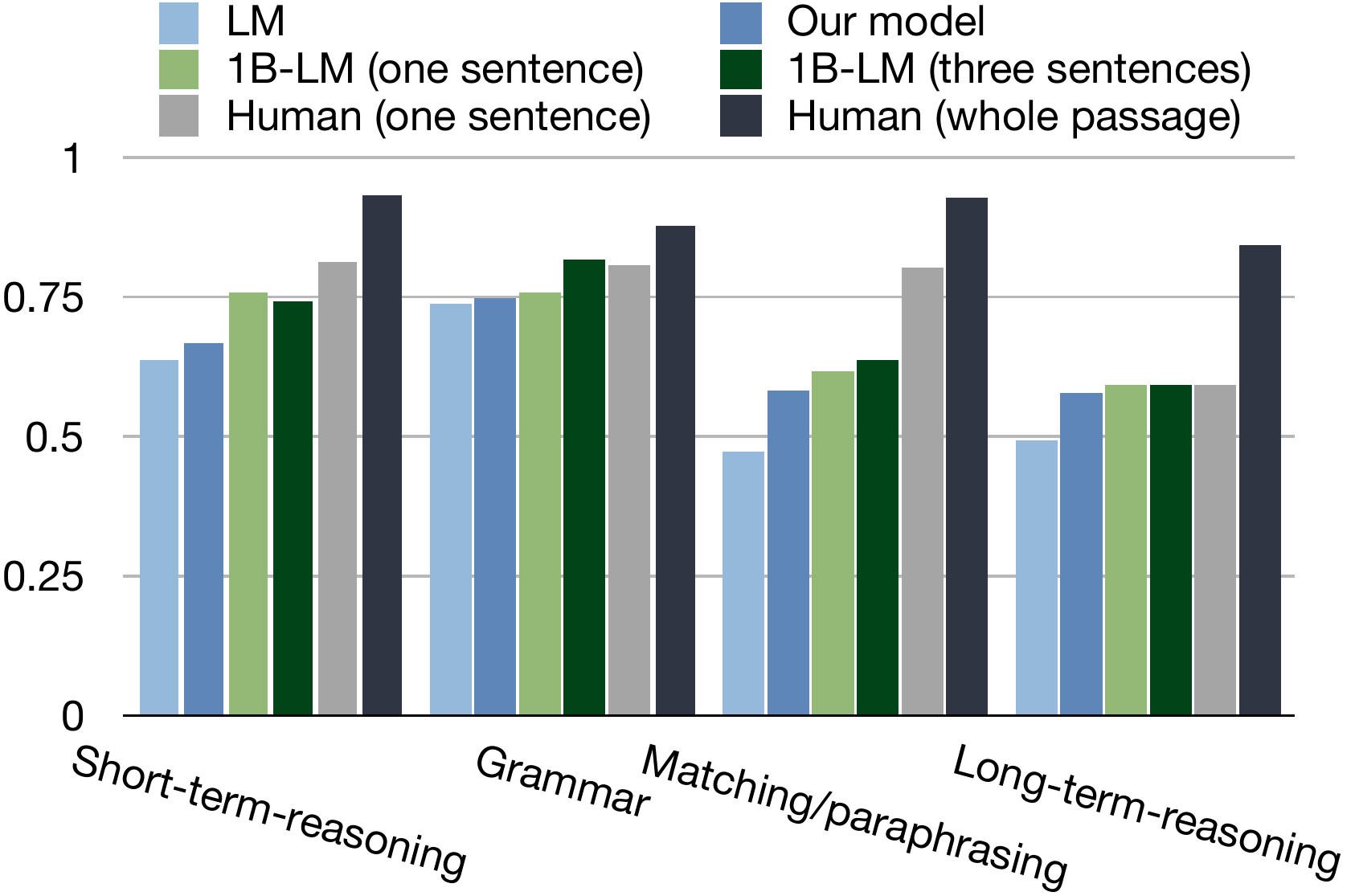}
        \caption{Middle school group (\namem)}
    \end{subfigure}%
    ~ 
    \begin{subfigure}[t]{0.5\textwidth}
        \centering
        \includegraphics[scale=0.4]{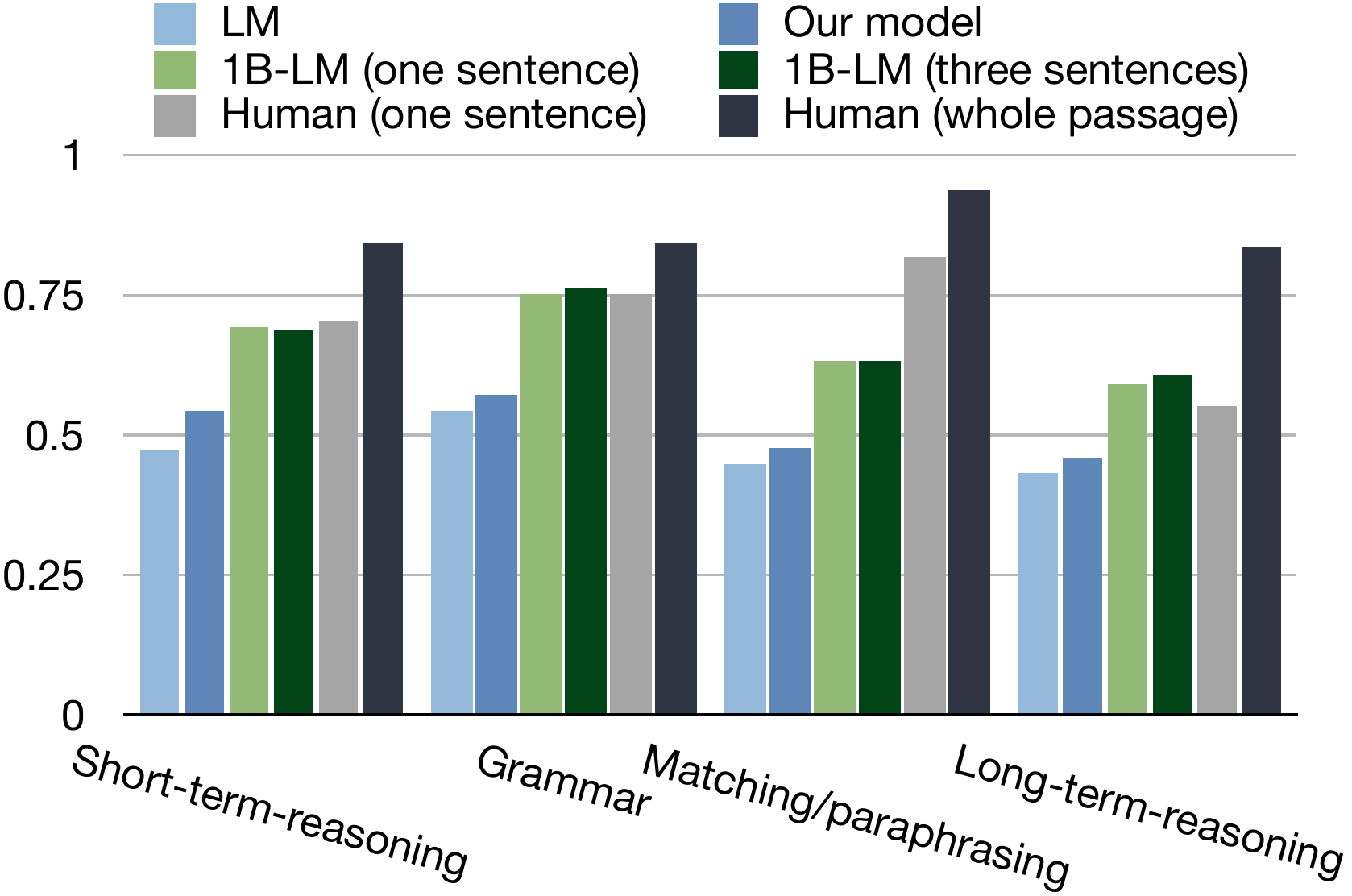}
        \caption{High school group (\nameh)}
    \end{subfigure}
    \caption{Model and human's performance on questions with different types. Our model will be introduced in Sec. \ref{sec:combine}.  
   }
    \label{fig:perform_sub_group}
\end{figure*}
\subsection{Type-specific Performance Analysis}
\label{sec:question_type_difficulty}
We can also  further verify the strengths and weaknesses of the 1B-LM by studying the performance of models and human on different question categories. Note that the performance presented here may be subject to a high variance due to the limited number of samples in each category.
From the comparison shown in Figure \ref{fig:perform_sub_group}, we see that 1B-LM is indeed good at short-term questions. Specifically, when the human only has access to the context of one sentence, 1B-LM is close to human's performance on almost all categories. 
Further, comparing LM and 1B-LM, we find that training on the large corpus leads to improvements on all categories, showing that training on a large amount of data leads to a substantial improvement in learning complex language regularities. 

\subsection{Implementation Details}
\label{sec:implementation}
We implement our models using PyTorch~\citep{paszke2017automatic}. We train our model on all questions in \name and test it on \namem and \nameh separately. For our final model, we use Adam~\citep{kingma2014adam} with the learning rate of $0.001$. The hidden dimension is set to $650$ and we initialize the word embedding by $300$-dimensional Glove word vector~\citep{pennington2014glove}. The temperature $\alpha$ is set to $2$. We tried to increase the dimensionality of the model but do not observe performance improvement.

When we train the small LM on \name, we largely follow the recommended hyperparameters in the Pytorch LM example\footnote{https://github.com/pytorch/examples/tree/master/word\_language\_model}. Specifically, we employ a $2$-layer LSTM with hidden dimension as $1024$. The input embedding and output weight matrix are tied.  We set the dropout rate to $0.5$. The initial learning rate is set to $10$ and divided by $4$ whenever the PPL stops improving on the dev set. 

We predict the answer for each blank independently for all of the models mentioned in this paper, since we do not observe significant performance improvements in our preliminary experiments when an auto-regressive approach is employed, i.e., when we fill all previous blanks with predicted answers. We hypothesize that, regardless of whether there exist inter-blank dependencies, since blanks are usually distributed far away from each other, LSTM is not able to capture such long dependencies. When testing language models, we use the longest text spans that do not contain blanks.

\begin{table}[th]

\begin{center}
\begin{minipage}[t]{350pt}
{\footnotesize
{\bf Passage:} 
Nancy had just got a job as a secretary in a company. Monday was the first day she went to work, so she was very \_1\_ and arrived early. She \_2\_ the door open and found nobody there. "I am the \_3\_ to arrive." She thought and came to her desk. She was surprised to find a bunch of \_4\_ on it. They were fresh. She \_5\_ them and they were sweet. She looked around for a \_6\_ to put them in. "Somebody has sent me flowers the very first day!" she thought \_7\_ . " But who could it be?" she began to \_8\_ . The day passed quickly and Nancy did everything with \_9\_ interest. For the following days of the \_10\_ , the first thing Nancy did was to change water for the followers and then set about her work. 

Then came another Monday. \_11\_ she came near her desk she was overjoyed to see a(n) \_12\_ bunch of flowers there. She quickly put them in the vase, \_13\_ the old ones. The same thing happened again the next Monday. Nancy began to think of ways to find out the \_14\_ . On Tuesday afternoon, she was sent to hand in a plan to the \_15\_ . She waited for his directives at his secretary's \_16\_ . She happened to see on the desk a half-opened notebook, which \_17\_ : "In order to keep the secretaries in high spirits, the company has decided that every Monday morning a bunch of fresh flowers should be put on each secretary’s desk." Later, she was told that their general manager was a business management psychologist.

\vspace{2ex}
\begin{tabular}{lllllc}
\multicolumn{5}{c}{{\bf Questions }}& {\bf Question type }\\
1. & A. depressed&B. encouraged&\textbf{C. excited}&D. surprised & short-term reasoning\\
2. & A. turned&\textbf{B. pushed}&C. knocked&D. forced &short-term reasoning\\
3. & A. last&B. second&C. third&\textbf{D. first} & long-term reasoning\\
4. & A. keys&B. grapes&\textbf{C. flowers}&D. bananas & matching\\
5. & \textbf{A. smelled}&B. ate&C. took&D. held & short-term reasoning\\
6. & \textbf{A. vase}&B. room&C. glass&D. bottle & long-term reasoning \\
7. & A. angrily&B. quietly&C. strangely&\textbf{D. happily } & short-term reasoning\\
8. & A. seek&\textbf{B. wonder}&C. work&D. ask & long-term reasoning\\
9. & A. low&B. little&\textbf{C. great}&D. general & long-term reasoning\\
10. & A. month&B. period&C. year&\textbf{D. week} & long-term reasoning\\
11. & A. Unless&\textbf{B. When}&C. Since&D. Before & grammar\\
12. & A. old&B. red&C. blue&\textbf{D. new} & long-term reasoning \\
13. & A. covering&B. demanding&\textbf{C. replacing}&D. forbidding  & long-term reasoning\\
14. & \textbf{A. sender}&B. receiver&C. secretary&D. waiter & long-term reasoning\\
15. & A. assistant&B. colleague&C. employee&\textbf{D. manager} & matching\\
16. & A. notebook&\textbf{B. desk}&C. office&D. house & matching\\
17. &\textbf{A. said}&B. written&C. printed&D. signed & grammar\\
\end{tabular}

\begin{tabular}{lllll}
\end{tabular}
}

\caption{An Amazon Turker's label for the sample passage}
\label{tab:sample_label}
\end{minipage}
\end{center}
\end{table}

\end{document}